\documentclass{llncs}

\usepackage[a-2u]{pdfx}
\usepackage[utf8]{inputenc}
\usepackage[T1]{fontenc}
\usepackage{xcolor}

\usepackage[breakable, theorems, skins]{tcolorbox}
\usepackage{amsmath}
\tcbset{enhanced}

\usepackage{microtype}
\usepackage{listings}
\definecolor{commentcolor}{rgb}{0.5,0.5,0.5}
\definecolor{darkgreen}{rgb}{0.09, 0.45, 0.27}
\definecolor{bgcolor}{rgb}{0.99,0.99,0.99}
\lstdefinestyle{rules}{
    morekeywords={if,else,not,rule,guard,action,allow,pred,isAboveThreshold,isBelowThreshold,hasRightValue1D,hasRightValue2D,hasRightCategories},
    basicstyle=\linespread{0.8}\sffamily\small,        
    breakatwhitespace=true,           
    showstringspaces=false,
    columns=fullflexible,
    tabsize=2,%
    breaklines=true,
    captionpos=b,
    numbers=left,
    numbersep=5pt,
    numberstyle=\sffamily\scriptsize\lsstyle,
    xleftmargin=0.05\columnwidth,
    commentstyle=\color{commentcolor},
    keywordstyle=\color{blue},
    stringstyle=\color{darkgreen},
    moredelim=**[is][\color{blue}]{@}{@},
    moredelim=**[is][\sl]{!}{!},
}
\lstdefinestyle{pseudocode}{
    basicstyle=\linespread{0.8}\sffamily\small,        
    breakatwhitespace=true,           
    showstringspaces=false,
    columns=fullflexible,
    tabsize=2,%
    breaklines=true,
    captionpos=b,
    numbers=left,
    numbersep=5pt,
    numberstyle=\linespread{0.8}\sffamily\tiny\lsstyle,
    xleftmargin=0.05\columnwidth,
    commentstyle=\color{commentcolor},
    keywordstyle=\color{blue},
    keywordstyle=[2]\color{red},
    stringstyle=\color{darkgreen},
    keywords={component,field,input,output,guard,condition,estimator,if,then,regression,time,to,type,ensemble,static,role,dynamic,with,cardinality,membership,T,in,unlimited,action,utility,priority,and,or,not,enter,n,situation,now,allow,value,estimate,where,true,each,notify,inner,isin,from,@},
    morecomment=[l]{\#},
    morekeywords=[2]{exclusiveSelect}
}
\newcommand{\longlstinline}[1]{{\small\textsf{#1}}}
\usepackage{subcaption}
\usepackage{footmisc}
\usepackage{paralist}
\usepackage{tabularx,colortbl}
\usepackage{array}

\usepackage{amsmath}

\usepackage{chngcntr}

\usepackage{fancyhdr}
\fancypagestyle{plain}{%
\fancyhead{}
\fancyfoot{}

\fancyfoot[L]{\scriptsize This is the authors' version of the paper: \textit{M. Töpfer, M. Abdullah, T. Bureš, P. Hnětynka, M. Kruliš: Ensemble-Based Modeling Abstractions for Modern Self-optimizing Systems, in Proceedings of ISOLA 2022, Rhodes, Greece, pp. 318-334, 2022}.\\ The final authenticated publication is available online at \url{https://doi.org/10.1007/978-3-031-19759-8_20}}
}
\fancypagestyle{empty}{%
\fancyhead{}
\fancyfoot{}

\fancyfoot[L]{\scriptsize This is the authors' version of the paper: \textit{M. Töpfer, M. Abdullah, T. Bureš, P. Hnětynka, M. Kruliš: Ensemble-Based Modeling Abstractions for Modern Self-optimizing Systems, in Proceedings of ISOLA 2022, Rhodes, Greece, pp. 318-334, 2022}.\\ The final authenticated publication is available online at \url{https://doi.org/10.1007/978-3-031-19759-8_20}}
}

\begin{document}
\clearpage
\pagestyle{plain}

\counterwithout{lstlisting}{chapter}

\title{Ensemble-based modeling abstractions for modern self-optimizing systems}

\author{
Michal Töpfer, 
Milad Abdullah, 
Tomas Bureš, 
Petr Hnětynka, 
Martin Kruliš
}
\institute{Charles University, Czech Republic\\ \email{\{topfer,abdullah,bures,hnetynka,krulis\}@d3s.mff.cuni.cz}
}
\authorrunning{Michal Töpfer et al.}

\maketitle

\begin{abstract}
In this paper, we extend our ensemble-based component model DEECo with the capability to use machine-learning and optimization heuristics in establishing and reconfiguration of autonomic component ensembles. We show how to capture these concepts on the model level and give an example of how such a model can be beneficially used for modeling access-control related problem in the Industry 4.0 settings. We argue that incorporating machine-learning and optimization heuristics is a key feature for modern smart systems which are to learn over the time and optimize their behavior at runtime to deal with uncertainty in their environment.
\end{abstract}


\section{Introduction}
\label{sec:introduction}
Modern smart systems increasingly aim at providing highly optimized behavior that copes with the high uncertainty and variability of their surrounding environment. This classifies them as representatives of \textit{self-adaptive systems}, which are systems that monitor their performance and their ability to meet their goals and adapt their behavior to cope with changes in themselves and in their environment.

Modern smart systems are often composed of multiple components, which are required to coordinate their operation. This means that the system adaptation and optimization have to be done in a coordinated manner too.

Traditionally these systems have been designed by providing a set of adaptation rules that were supposed to identify principal states and changes in the environment and reconfigure the system. However, with more and more emphasis on data and with the increasing amounts of data the modern systems collect and are able to exploit in their operation, the specification using a fixed set of adaptation rules becomes increasingly more complex and hard to do (mainly due to the fact that the model behind the data is unknown and potentially changing).

A recent trend in modern smart systems is to use machine learning to help the system make adaptation and optimization decisions. However, machine learning is still used in a rather ad-hoc manner without having been systematically embedded in the architecture of the systems. 

The main problem is that there is a lack of architectural models for the specification of system components that would provide abstractions for machine learning and other kinds of optimizations based on data observed by the system. As a result, every smart system that employs machine learning or some kind of optimization has to create its own architectural abstractions. This not only means duplicating the work but also requiring an expert in machine learning to be able to implement the abstractions correctly.

In this paper, we address this problem by providing a novel component model (called ML-DEECo) that features dedicated abstractions for machine learning (currently only supervised learning) and optimization heuristics (used mainly for coordination problems).

The ML-DEECo model is an extension of our previous DEECo component model~\cite{bures_deeco_2013}. Similar to DEECo, it is based on the concept of autonomic component ensembles, which are situation-based coordination groups of components that cooperate (within a given situation) to achieve a common goal.

In this paper, we demonstrate the main concepts of ML-DEECo on a use-case in the Industry 4.0 settings that stems from our previous project with industry.

We support the concepts of ML-DEECo by an implementation in Python, which can process the ML-DEECo abstractions and use them to automatically realize the complete learning loop (consisting of data collection, model training, and inference at runtime). By this, we show the usefulness of ML-DEECo as it saves not only the time needed to develop the learning loop but also the expertise needed to do that.
The implementation is part of the replication package that comes along with the paper~\cite{replication}.

The structure of the text is as follows.
Section~\ref{sec:example} describes the running example and, based on the example, introduces ensemble-based component systems.
Section~\ref{sec:estimates} presents the machine-learning concepts of ML-DEECo while Section~\ref{sec:heuristics} discusses the application of heuristics.
Section~\ref{sec:evaluation} evaluates the presented approach and Section~\ref{sec:related} compares it with related work.
Section~\ref{sec:conclusion} concludes the paper.

\section{Running example and background}
\label{sec:example}
In this paper, we build on the running example from recent work on dynamic security rules by Al-Ali et al.~\cite{Al_Ali_2019}. 
The example is a simplified version of a real-life scenario concerning access to a smart factory and it is taken from our recent project with industrial partners.

The factory consists of several working places, each with an assigned shift of workers. The workers come to the factory in the morning. They have to go through the main gate to enter the factory. Then, they have to grab a protective headgear at a headgear dispenser inside the factory and they can continue to their workplace. The workers are not allowed to enter the workplaces of the other shifts in the factory.

To ensure security and safety in the factory, several rules are defined. For example, the workers are allowed to enter the factory at the earliest 30 minutes before their shift starts. The rule is enforced by a smart lock at the main gate. Another rule defines that the worker can only enter his workplace if they are wearing the protective headgear. We can see that the security rules are dynamic as their effect depends on time (we allow workers to the factory at most 30 minutes before their shift) and on of workers (whether they wear the headgear or not).

The scenario further deals with replacing the workers who are late for their shift with standbys. For each shift in the factory, we have a defined set of standby workers. If a worker does not arrive at the factory sufficiently (16 minutes) before their shift starts, they are canceled for the day and a standby is called to replace them. As the actual time of arrival of the workers to the factory varies, it brings uncertainty to the system, which is an opportunity to use a machine-learning-based estimate to deal with it. Furthermore, the assignment of the standbys can be a hard optimization problem if the sets of standbys for the shifts overlap, so it can be dealt with using a heuristic approach.

\subsection{Modeling dynamic security rules with components and ensembles}

We build our approach on the concept of autonomic component \emph{ensembles}; in particular, we are using the abstractions of the DEECo ensemble-based component model~\cite{bures_language_2020}. In this section, we describe the concepts of ensembles that are necessary for understanding the paper. 

Components represent the entities in our system (factory, rooms, dispenser, workers) and they are described by the \lstinline{component type}.
There are multiple instances (simply referred to as “components” in the paper) of each of the component types (i.e., multiple workers, rooms, etc.).
Each of the components has a state represented as a set of data fields.
Components that we cannot control directly (workers in our case) are referred to as “beyond control components”. 
The state of these components is only observed or inferred.

Ensembles represent dynamically formed groups of components that correspond to a coordinated activity. 
As for the components, the ensembles are described by their \lstinline{ensemble type}, which can be instantiated multiple times.
A single component can be selected in multiple ensemble instances at the same time---we call this that the component is a “member” of the respective ensemble instance. 
The ensemble prescribes possible data interchange among the components in the ensemble and also the group-wise behavior (e.g., a coordinated movement).

Technically, an ensemble type specifies the (1) roles, (2) situation, and (3) actions.

The \textit{roles} determine which components are to be included in the ensemble (i.e., components that should be members of the ensemble). 
Roles are either static or dynamic. 
The static ones are specified when the ensemble is instantiated and cannot change. They provide the identity to the ensemble instance (e.g., a shift for which the ensemble is instantiated). There cannot exist two ensemble instances of the same type that are equal in the assignment of their static roles.

The static roles are specified as a tuple: (i) type of component instances in the role, (ii) cardinality of the role.

The cardinality of the role is simply an interval determining the minimum and maximum number of component instances referred to by the role.

On the other hand, the dynamic roles are populated dynamically based on the situation in the system. 
The dynamic roles are specified as a triple: (i) type of component instances in the role, (ii) cardinality of the role, and (iii) condition over components in the role. 

While the type and cardinality are the same as in the case of static roles, the condition further determines if a component may be selected for the role. The condition is parameterized by the component being selected for the dynamic roles and by the static roles (i.e. attributes of components in the static roles).

The situation is a condition over static roles that determines whether the ensemble should be instantiated.

The actions are activities to be executed when the ensemble is instantiated. They differ based on the target domain and have no influence on how ensembles are instantiated. In the case of our running example, we use two actions: \textit{allow} and \textit{notify}. The \textit{allow} action assign some permission to a component. The \textit{notify} action send a notification to a component (e.g. to let a standby worker know that they have been assigned to a shift).

Ensembles can be hierarchically nested, i.e., an ensemble can be defined as an inner one within another ensemble.
This allows for the decomposition of complex conditions into simpler ones and thus makes the whole specification more easily manageable.
Instances of an inner ensemble can be created only if there is an instance of the outer ensemble.

The ensembles are instantiated and dissolved dynamically and continuously as the state of the system changes. This instantiation of the ensembles and their dissolution is performed automatically by the ML-DEECo runtime framework.

An ensemble is instantiated for each possible assignment of static roles such that the situation condition is true and there exist components that can be assigned to the dynamic roles to satisfy the cardinality bounds and the dynamic role condition. 

The instantiation is attempted periodically. In case of our running example, since this is run as a simulation, we perform the instantiation (and dissolution) of ensembles in every simulation step.

For our example, ensembles are used to express the security rules in the system. They dynamically grant/revoke access permissions to individual components.

\begin{figure}[h]
    \centering
    \includegraphics[width=\columnwidth]{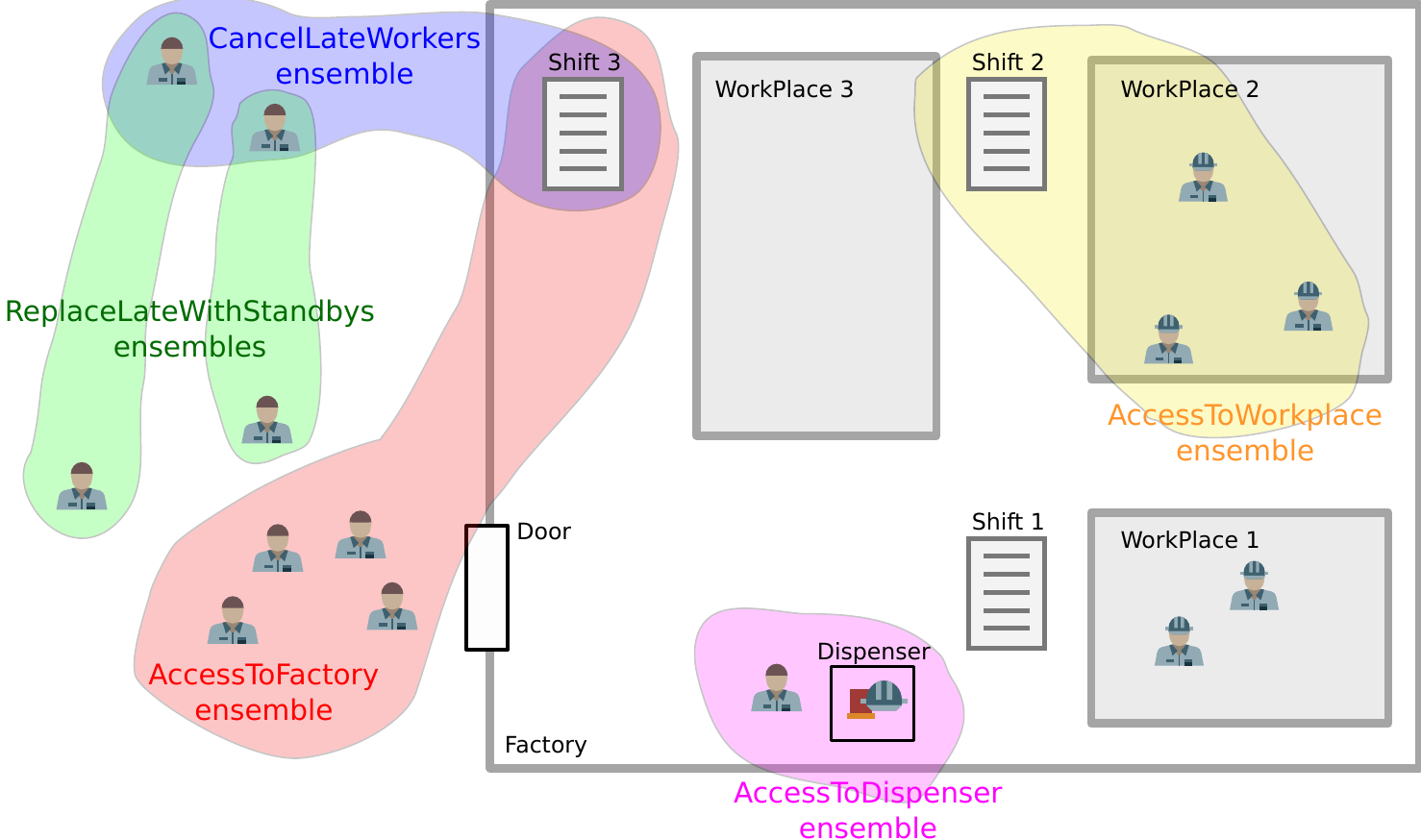}
    \caption{Factory example}
    \label{fig:factory}
\end{figure}

Listing~\ref{lst:access_to_factory} shows a part of the specification of the running example. It represents the components and ensembles as illustrated in Figure~\ref{fig:factory}
There are 6 types of components: \lstinline{Door}, \lstinline{Dispenser} (of protective headgear), \lstinline{Factory}, \lstinline{WorkPlace}, \lstinline{Shift} and \lstinline{Worker}. 
Each of them is described by a set of their fields.

\begin{lstlisting}[escapechar=|,breaklines=true,style=pseudocode,label=lst:access_to_factory,caption=An excerpt of the example specification]
component type Door:
  field position: Position
  
component type Dispenser:
  field position: Position

component type WorkPlace
  field position: Position

component type Factory
  field entryDoor: Door
  field dispenser: Dispenser
  field workplaces[*]: Workplace

component type Worker
  field position: Position
  field hasHeadgear: boolean

component Shift
  field workPlace: WorkPlace
  field startTime: Time
  field endTime: Time
  field assigned[*]: Worker  # an original list of assigned workers
  field workers[*]: Worker   # a list of actually working workers
  field standBys[*]: Worker

ensemble type AccessToFactory|\label{inlst:ens-start}|
  static role shift: Shift|\label{inlst:access_to_factory:static_role}|
  situation shift.startTime - 30 <= now <= shift.endTime + 30|\label{inlst:access_to_factory:situation}|
  dynamic role workers[*]: Worker|\label{inlst:access_to_factory:dynamic_role}|
    each worker in workers:
      (worker in shift.assigned and not worker.canceled)|\label{inlst:access_to_factory:membership:beg}|
        or worker in shift.calledStandbys|\label{inlst:access_to_factory:membership:end}|
  action allow workers enter factory|\label{inlst:access_to_factory:allow}|
  
ensemble type CancelLateWorkers|\label{inlst:ens-cancel-late}|
  static role shift: Shift
  situation shift.startTime - 16 <= now <= shift.endTime
  dynamic role lateWorkers[*]: Worker
    each worker in lateWorkers:
      worker in shift.assigned and not worker.isAtFactory 
  action notify lateWorkers with canceled|\label{inlst:cancel-worker}|
  inner ensemble type ReplaceLateWithStandbys|\label{inlst:ens-replace}|
    dynamic role standBys[lateWorkers.size]: Worker
       each worker in standBys:
         worker in shift.standBys 
    action notify standBys with calledIn
\end{lstlisting}

Further, ensemble types are specified.
Due to the paper space limits, we show here only a few of them. In Figure~\ref{fig:factory}, sample instances of ensembles are represented as the hand-drawn oval shapes.

The \lstinline{AccessToFactory} assigns to the workers the access-to-factory permission.
The static role \lstinline{shift} (line~\ref{inlst:access_to_factory:static_role}) determines the identity of the ensemble instance, i.e., the ensemble is instantiated for a selected shift and there is only one ensemble instance of \lstinline{AccessToFactory} for a particular shift.

The \lstinline{situation} (line~\ref{inlst:access_to_factory:situation}) is a predicate determining under which conditions the ensemble has to be instantiated.
In this particular case, the ensemble is instantiated from 30 minutes before the shift starts till 30 minutes after the shift ends.

The dynamic role \lstinline{workers} (line~\ref{inlst:access_to_factory:dynamic_role}) selects all the workers which are assigned to the shift. These are workers that have been assigned to the shift and that have not been canceled due to being late, plus the standbys called to fill in for the late workers. This is stated on lines~\ref{inlst:access_to_factory:membership:beg}--\ref{inlst:access_to_factory:membership:end}. The definition of the role specifies the cardinality, the type of components assigned to the role, and the condition that has to hold for each component in the role.

The workers selected in the dynamic role are assigned with permission to enter the factory (line~\ref{inlst:access_to_factory:allow}). 

Similar to the \lstinline{AccessToFactory}, our specification contains ensembles \longlstinline{AccessToDispenser} and \lstinline{AccessToWorkplace}, which differ only in what they regulate the access to. Since they are very similar, we omitted them in the paper.

The next ensemble---\lstinline{CancelLateWorker} (line \ref{inlst:ens-cancel-late}) serves for canceling workers that are late.
The static role pins the ensemble instance to a shift. The dynamic role selects workers that are late (i.e., they are not in the factory 16 minutes before the shift start).
Those workers are then canceled from the shift (line \ref{inlst:cancel-worker}). This is done by notifying them and the system about being canceled.

To replace the canceled workers, there is a subensemble (\longlstinline{ReplaceLateWithStandbys} in line~\ref{inlst:ens-replace}) of the \lstinline{CancelLateWorker} ensemble. It selects the necessary number of standbys to fill in for the canceled workers. The cardinality of the \lstinline{standBys} role is equal to the number of canceled workers.
The standbys are notified by the ensemble about being called in.

All the described ensembles are illustrated in Figure~\ref{fig:factory}.
The figure also shows other two ensembles (omitted from the listing due to the space constraints but defined in the complete example in the replication package).
They are \lstinline{AccessToWorkspace} assigning permissions to access the particular workplace and \lstinline{AccessToDispenser} assigning permission to obtain the headgear.

\section{Estimates}
\label{sec:estimates}
In this section, we present the extension of the DEECo component model which brings in  machine-learning-based estimates. We call the extended component model ML-DEECo. Though we demonstrate the concepts on the running example, our approach is general and can be used in other use cases as well. The ML-DEECo framework allows the architect of the system to include the estimates inside both components and ensembles and to use the estimated values for adaptation decisions. The Python implementation, together with a more detailed description of all the provided abstractions, is available in the replication package~\cite{replication}.

\subsection{Estimates}

In our recent projects, we have identified several use-cases, where machine learning would be beneficial to a proactive self-adaptation of the system. Several of these use cases exhibited the same patterns---we needed to predict a future state of a component or ensemble. The machine learning algorithms can easily be used in such situations as the future state will be known at some point in time in the future so we can formulate the problem as supervised learning and use the observed value for the training of the machine-learning-based estimate.

Based on our previous experience with designing ensemble-based systems and an analysis, we have identified the following three kinds of estimates based on where they can be applied in an architecture, i.e., on
\begin{inparaenum}[(i)]
\item a component,
\item an ensemble, and
\item a component-ensemble pair.
\end{inparaenum}

In the first case, the estimate serves as a special field of the component and is parameterized by values of the regular fields.
For instance, the \lstinline{Worker} component can have an estimate predicting whether the particular worker would be willing to be activated as a ``standby'' (in the case the selection as a called-in standby is voluntary).
The estimate would be parameterized by the current day-of-week and would be trained on the history of the particular worker's willingness to be activated as the standby.

In the second case, i.e., applying on an ensemble, an estimate is employed as the ensemble property and can be used in the ensemble conditions (in the situation and the dynamic role selectors).
The estimate is parameterized by values obtained from global variables and static roles of the ensemble. For instance, if we extend the example by the automated planning of shifts (i.e., allocating workers to shifts), we can have an ensemble with the estimate that predicts the right number of workers in order to reach expected productivity goals. This prediction would be based on the day-of-week and start time of the shift (morning, afternoon, evening).

The last option---applying an estimate to the component-ensemble pair---is the most complex one. In this case, we associate the estimate with a component that has been dynamically selected in an ensemble. Since this is the most complex case, we will use it to showcase the ML-DEECo in the rest of the section.

As an example, we use machine learning for the adaptation of the replacement of late workers. A traditional rigid solution is to set a threshold for when the workers have to be inside the factory. In our case, the baseline approach is to cancel the workers if they are not present in the factory 16 minutes before the start of their shift. This, however, does not deal well with the uncertainty of the arrival of the workers. To replace the rigid approach, we suggest using a machine-learning-based estimate to predict whether a worker will come on time or not. This can be expressed as an estimate assigned to a dynamic role of an ensemble type as shown in Listing~\ref{lst:cancel_late_workers}.

\begin{lstlisting}[escapechar=|,breaklines=true,style=pseudocode,label=lst:cancel_late_workers,caption=Machine-learning-enabled ensemble specification -- Cancel late workers.]
ensemble type CancelLateWorkers
  static role shift: Shift|\label{inlst:cancel_late_workers:static_role}|
  situation shift.startTime - 30 <= now <= shift.endTime + 30|\label{inlst:cancel_late_workers:situation}|
  dynamic role lateWorkers[*]: Worker|\label{inlst:cancel_late_workers:dynamic_role}|
    with value estimate willArrive:|\label{inlst:cancel_late_workers:estimate}|
      output worker.isAtFactory @ T+<1,30>|\label{inlst:cancel_late_workers:output}|
      input dayOfWeek|\label{inlst:cancel_late_workers:input}|
      guard worker.shift == shift|\label{inlst:cancel_late_workers:guard}|
    each worker in lateWorkers
       not worker.isAtFactory and not (willArrive @ shift.startTime) |\label{inlst:cancel_late_workers:membership}|
  action notify lateWorkers with cancelled|\label{inlst:cancel_late_workers:action}|
\end{lstlisting}

The dynamic role for finding late workers is specified on line~\ref{inlst:cancel_late_workers:dynamic_role}. We assign an estimate to this role on line~\ref{inlst:cancel_late_workers:estimate} to predict whether the worker will come to the factory or not. We use the prediction to determine whether we should cancel the worker or not. 
That is done on line~\ref{inlst:cancel_late_workers:membership} in the membership condition of the role -- we use the estimate to predict whether the worker will be at the factory at the time the shift starts.

To be able to use the estimate, we need to specify its inputs and outputs. The output of the estimate is the future value of the \lstinline{worker.isAtFactory} attribute (line~\ref{inlst:cancel_late_workers:output}). By \lstinline{T+<1,30>}, we indicate that we want to construct a model which is able to predict this value in a range of 1 to 30 minutes into the future. The input of the model is the day of the week which the shift takes place on (line~\ref{inlst:cancel_late_workers:input}). Furthermore, we define a \lstinline{guard} on line~\ref{inlst:cancel_late_workers:guard} which indicates the validity of the training data -- here, as we have an instance of the ensemble for each shift, we only want to collect data about the workers from the one shift and not the other shifts.

\subsection{Training of the estimates}

The specification of the estimate in the ensemble definition is enough to be able to automatically collect data and train the machine learning model, which is done by the ML-DEECo framework runtime. The semantics of the value estimate is as follows.

The estimate predicts an attribute from the future state of the system. 
In our example, the \lstinline{willArrive} estimate predicts a future value of the \lstinline{isAtFactory} attribute of the \lstinline{Worker} component.
We use the values of the attributes available at the time of prediction as inputs to the machine learning model---\lstinline{dayOfWeek} in the example. Furthermore, we need to specify how far into the future we want to be able to predict---this is set by the \lstinline{T+<min_t,max_t>} expression.
This will influence what data we collect for training---we allow the time difference between inputs and outputs to be in the range $\langle \textit{min\_t}, \textit{max\_t}\rangle$.

To train the machine learning model, we need to collect training data. Our focus is on predictions of values, which can be observed at some point in the future. In the example, we predict whether the worker will come or not, and after some time, we can observe whether they really came or not. We thus need to observe these values and use them for the training of the model.

To collect the training data, we use the following procedure. In every time step of the simulation (every minute in the example), we check the guard and if it marks the data as valid, we collect the values of input attributes and tag them with the current time. After that, we collect the values of the output attributes for the current time step and link them with past inputs. Specifically, we go through all the allowed time differences $t \in \langle \textit{min\_t}, \textit{max\_t}\rangle$ and save the training example \lstinline{(t, inputs_at_now-t, outputs_now)}. The inputs are collected throughout the whole run of the simulation and the training is performed after the simulation finishes.

\section{Heuristics}
\label{sec:heuristics}
As mentioned in Section~\ref{sec:example}, after canceling late workers, it is necessary to select replacements from the standby workers. As this is not a prediction problem, we do not use machine learning (which we presented in the previous section). Rather we look for abstractions that allow us to specify component partitioning.

The problem here is that the assignment of standby worker is mutually exclusive -- a standby worker cannot be assigned in place of two different late workers. This leads to an optimization problem, which has known inputs (i.e. components and their attributes, including the predicted ones) and known constraints. 

An optimal selection of the standbys, considering for instance shared standbys among shifts, is an NP-complete problem. We faced this issue extensively in our previous work~\cite{bures_language_2020}, where we were transforming ensembles to the constraint optimization problem.
Even for small problem instances, the constraint optimization problem would soon become too computationally expensive to solve. 

However, in this particular case, the optimal solution is not necessary (i.e., the correctness of the system here does not depend on whether the standbys are selected optimally but only that selections do not overlap---note that this is in strict contrast with the access permission assignment, where strict correctness is necessary).
Fortunately, there are a number of heuristic algorithms that target such problems (i.e., NP-hard or complete problems).
For example, a well-known and widely used heuristic algorithm is a k-means clustering method~\cite{kanungo_efficient_2002}, which partitions $n$ elements into $k$ clusters in which each element belongs to the cluster with the nearest mean.
By itself, the problem is NP-complete, but the k-means clustering method quickly converges to a local optimum.

In the case of the standbys selection, the exclusive choice method (e.g., \cite{heinrich_automated_2015}) can be used. 
Listing~\ref{lst:exclusive-choice} shows the updated version of the \lstinline{ReplaceLateWithStandbys} ensemble.
Now, the selection of standbys is performed with the help of a special operation (line~\ref{inlst:exclusive}), where the exclusive choice heuristic method is implemented.

\begin{lstlisting}[escapechar=|,breaklines=true,style=pseudocode,label=lst:exclusive-choice,caption=Ensemble specification with heuristic operation]
ensemble type CancelLateWorkers
  # ... shortened here
  inner ensemble type ReplaceLateWithStandbys
    dynamic role standBys[lateWorkers.size]: Worker
       exclusiveSelect worker from globalStandBys|\label{inlst:exclusive}|
    action notify standBys with calledIn
\end{lstlisting}

The above-mentioned k-means clustering method could be employed in our example, e.g., for clustering workers for optimized delivery of sensor data (i.e., workers are equipped with sensors, and to reduce the amount of communication, the measurements are not delivered directly, but they are aggregated by a worker that is "in-the-middle" of each cluster).

The downside of the use of heuristics is that they target individual problems.
Therefore it is complicated to create a common "heuristics extension" for ML-DEECo, and particular heuristics need to be offered as specialized operations. However, our experience shows that there is only a limited number of problems in ensemble-based systems related to the assignment of components to ensembles: selection (exclusively select a fixed number of components for each ensemble instance), partitioning (split all components in a set among a set of known ensemble instances), clustering (split all components to an optimal unknown number of ensembles). For these, abstractions like the \lstinline{exclusiveSelect} above can be created.

As a future work, we plan to create a detailed classification of problems related to ensemble-based systems and based on it, to design a set of common---at least to some extent---heuristics for ML-DEECo.

\section{Evaluation}
\label{sec:evaluation}
We center the evaluation around two arguments: (1) it saves on implementation effort to have machine learning abstractions part of the component model, (2) for certain situations allows the systems to perform better than just with fixed adaptation rules.

To this end, we have implemented a simulation of the running example using the ML-DEECo Python framework. 

From the perspective of implementation effort, when using the abstractions of ML-DEECo realized by our Python implementation, the introduction of machine learning consist essentially only in declaring the predictor and providing annotations to component fields and ensemble roles. The exact code can be found in the replication package~\cite{replication}. All this amounts to approx. 25 lines of code. On the other hand, custom implementation using TensorFlow or PyTorch frameworks would amount to at least a hundred lines of code (taking into account the data collection, pre-processing, training, retraining, and inference---all taken care of by ML-DEECo). 

In addition to the Industry 4.0 example presented in this paper, we have used ML-DEECo to model also scenarios from smart farming. In all cases, the abstractions featured by ML-DEECo proved to be expressive enough to cover various supervised learning situations---including predictions of future states of sensors (i.e., continuous data), prediction of the future state of a component (i.e., discrete data), predictions about the existence of an ensemble, and predictions about which components will be members of an ensemble.

To illustrate that machine learning has the potential to outperform fixed adaptation rules, we simulated the Industry 4.0 example described in the paper. The configuration of the simulation is as follows. We assume that the workers arrive by a bus which stops a few minutes from the main gate of the factory. During business days, the bus arrives 24 minutes before the shift starts, and during weekends, the bus arrives 30 minutes before the shift. Furthermore, we assume that 10\% of the workers are late each day and they arrive by a later bus---18 minutes before the start of the shift on business days and 15 minutes before the shift during weekends. To have more uncertainty in the environment, we add a random delay (with exponential distribution) to each worker. If a worker is canceled and replaced by a standby, we assume that it will take 30 minutes before the standby arrives. We chose these values after careful deliberation to have something that
both illustrate the system and are close to what the reality would look like.

\begin{figure}[h]
    \centering
    \includegraphics[width=\columnwidth]{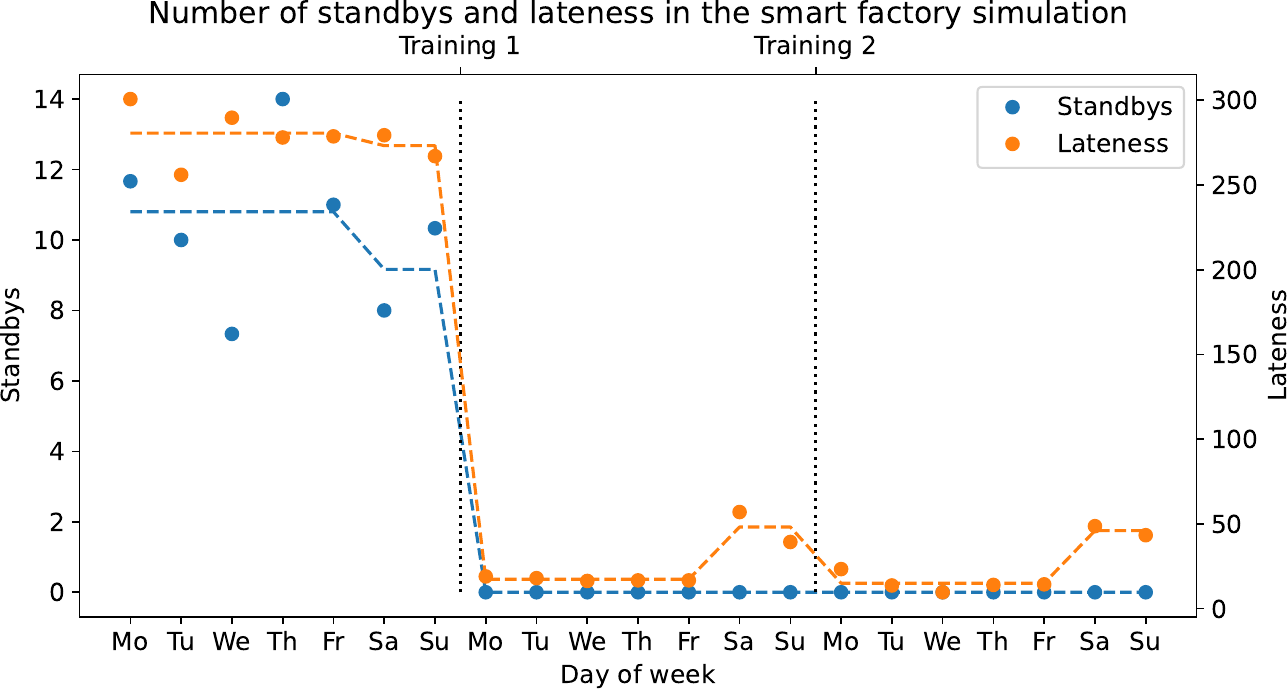}
    \caption{Results of the simulation with 100 workers.}
    \label{fig:results}
\end{figure}

We ran the simulation with three shifts starting at the same time, each with 100 workers, for three iterations. In the first iteration, we use the rigid rule of canceling workers 16 minutes before their shift starts. In the following two iterations, we use the machine-learning-based estimate described in Section~\ref{sec:estimates} to decide whether to cancel the worker. The results are shown in Figure~\ref{fig:results}. The blue points are the number of necessary standbys (averaged over the shifts) with a blue line showing the average over the business days and weekends. The \textit{lateness} (shown in orange) is computed as the square of the delay of workers who arrive late at their workplace. The ML-based estimate is trained after each iteration (week)---first, we train it on the data collected while using the rigid rule (denoted Training 1 in the figure), then, we update it using data collected in the second week (Training 2). It is clear that for this configuration, the learned rules perform significantly better than the rigid rule.

\begin{figure}[h]
    \centering
    \includegraphics[width=\columnwidth]{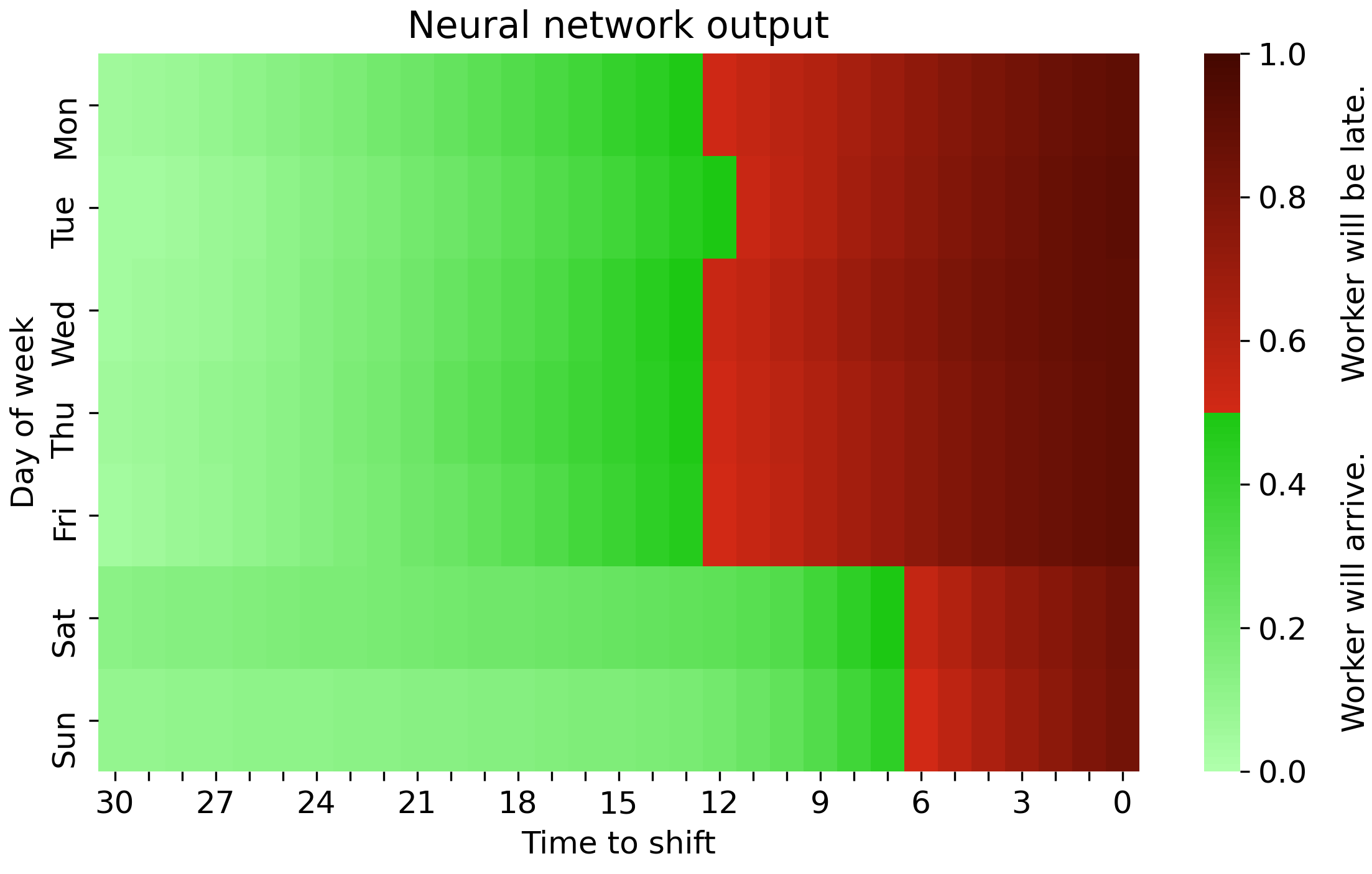}
    \caption{Outputs of the neural network for predicting whether a worker will arrive to the factory before their shift starts (green) or not (red).}
    \label{fig:nn}
\end{figure}

We have further inspected the outputs of the machine learning model to see what the learned threshold for cancellation is. The outputs are shown in Figure~\ref{fig:nn}. We observe that the learned threshold is different for business days and for weekends (as we want it to be). The model is more forgiving to the workers than the rigid rule---it allows them to enter the factory at latest 12 minutes before the shift starts on business days, and 7 minutes before the shift starts on weekends. It seems that it is beneficial to wait for the workers arriving by the later bus instead of calling the standbys which will take longer to arrive.

Furthermore, we have tried running the simulation with 20\% and 30\% of late workers to see the impact on the learning process. The results were very similar and the learned rule performed significantly better than the rigid one. For 30\% of late workers, the difference in the cancellation time for business days and for weekends was not that big, but it is still clear that the network was capable of learning the pattern. The plots with results for 20\% and 30\% of late workers are available in the replication package~\cite{replication}.

In the example, we focused on a single source of uncertainty---the arrival of the workers.
Other type of uncertainty in the example could be predicting whether a worker actually grabs the headgear. 
This is essentially an identical problem to the uncertainty of the workers arrival, only the predicted values are not continuous but discrete (and thus  the underlaying neural network would differ only in the last layer that would use softmax, and in the loss function that would use categorical cross-entropy). 
For the sake of conciseness, we focused on a single example only.

\textit{Limitations and Threats to Validity:} The evaluation we presented serves as an indication of an illustration of the potential of our approach. Due to the limited size of the use case, we refrain from claiming the generality of the approach. However, at least our indicative experiments show its potential.

We aimed at giving generality to the abstractions by constructing them independently of a particular use case. We based the abstractions on a taxonomy prediction (briefly discussed in Section~\ref{sec:estimates})---i.e., associating predictions with a component, ensemble, or component-ensemble pair and classifying types of predictions. This is independent of our use case and has been built as a combination of generally accepted abstractions.

Our work is limited to supervised learning only. We are currently in the process of extending ML-DEECO of reinforcement learning.

\section{Related Work}
\label{sec:related}
In the paper, we have presented a component model targeting the development of adaptive systems and employing machine learning techniques directly at the level of a system architecture specification.
Thus, the related areas are approaches defining explicit component models for adaptive systems and approaches combining adaptive systems with machine learning techniques---ideally, approaches combining both.

Using machine-learning techniques in adaptive systems is not a new approach. 
In~\cite{saputri_application_2020}, a systematic literature review (SLR) analyzes the employment of machine learning techniques in adaptive systems for the past 20 years.
From the analysis, an apparent increasing trend in their usage can be seen, and the machine learning techniques are mainly employed in the adaptation phase.
Also from the same area is the SLR in~\cite{gheibi_applying_2021} which also confirms the increasing usage of machine learning techniques in adaptive systems---again mainly in the adaptation phase to optimize it and/or predict future actions.
Nevertheless, most of the analyzed approaches use machine learning techniques ``under-the-hood'' in their implementation.
This is in strict contrast with our approach, where we explicitly use and expose them for the architecture specification.
In the text below, we discuss selected related approaches in more detail.

There are different uses for machine learning in self-adaptive systems ranging from predicting sensor values to reducing the space of adaptations. 
E.g., in~\cite{van_der_donckt_applying_2020} neural networks are employed for such a reduction.
A similar approach~\cite{damiani_effective_2019} (by a similar group of authors) combines machine-learning techniques with a cost-based analysis to reduce the space and choose an adaptation with the best cost.
In~\cite{gheibi_impact_2021}, a theorem defining a theoretical bound on the impact of applying a machine-learning method during adaptation was defined, and an approach for reducing an adaptation space was proposed.
The paper~\cite{gabor_scenario_2020} proposes a framework for the coevolution of an adaptive system together with its tests.
Machine-learning is used for restrictions of an adaptation space in order to achieve a meaningful system as a result of an adaption step.
An approach of a combination of machine-learning techniques and probabilistic model checking is described in the paper~\cite{camara_quantitative_2020} and used for choosing the best adaptation and refusing unfeasible ones.
The approach thus allows for fast convergence towards optimal decisions.
Online reinforcement learning is used in~\cite{palm_online_2020}  to deal with design-time uncertainty and automation of the development of the self-adaptation logic.
Thanks to automation, there is thus no need for manual activities during the application of reinforcement learning.
In the SARDE framework~\cite{grohmann_sarde_2021}, machine learning is used for the selection of the best estimation approach and for optimization of the selected approaches.
The whole framework then allows for self-adaptive resource demand estimation.
In~\cite{muccini_machine_2019}, machine-learning is utilized for forecasting values of QoS parameters and, therefore, for advanced and proactive selection of possible adaptation strategies.

As mentioned, the approaches in the paragraph above use machine-learning techniques internally for a particular functionality of a system/framework.
This is in contrast with our approach, where we are targeting the creation of architectural abstractions allowing for the usage of machine-learning during the design of a self-adaptive system and direct use of machine-learning results at the architectural level. 

This leads to the second area of related works---component models for adaptive systems and especially those that offer implementation in a common programming language.
The Service Component Ensemble Language (SCEL)~\cite{nicola_scel_2015} laid the mathematical foundations and semantics for ensemble-based systems.
Later, the concepts of SCEL were implemented in a Java-based runtime framework called jRESP~\cite{jresp}.
Another implementation of the ensemble-based concepts can be found in Helena, which is a complete framework for developing ensemble-based systems~\cite{hennicker_foundations_2014}. An approach similar to ensembles can be found in the $AbC$ calculus~\cite{alrahman_power_2016} that defines systems via attribute-based communication between components.
The calculus has been formalized in~\cite{alrahman_programming_2020}. 
Implementation of $AbC$ is available as the $Ab^aCuS$ framework~\cite{alrahman_programming_2016}. 
Similar to our implementation of ML-DEECo, components in $Ab^aCuS$ are modeled as classes, and processes (performing communication via component attributes similarly to ensembles) are also classes.                                           
Another implementation of $AbC$ is ABEL~\cite{riis_nielson_abel_2019} that is a DSL developed in the Erlang language.
Dynamic logic for describing ensemble-based communication between components is defined in~\cite{hennicker_dynamic_2020}, and it is implemented in a variant of the $AbC$ calculus.
The DReAM framework~\cite{de_nicola_dream_2020} allows for specifications of dynamically reconfigurable architectures. 
Its architecture description language is based on an interaction logic and describes dynamic coordination among components. 
Static parts of the architecture are described using a propositional interaction logic, while DReAM describes dynamic coordination. 
The Java implementation of DReAM, similarly to our approach, maps components and coordination to classes.
The BIP component model~\cite{bliudze_algebra_2008} is used as a basis for the propositional interaction logic employed in DReAM.
BIP primarily focuses on the formal description of component behavior. 
A combination of UML components and BIP is proposed in~\cite{chehida_component-based_2021}, and it focuses on the description and verification of component behavior and inter-component communication. 
An extension of BIP is DR-BIP~\cite{el_ballouli_programming_2021} that adds support for dynamic reconfigurations.

The approaches for modeling and implementing ensemble-based (and similar) architecture primarily focus on semantics but they do not introduce any machine-learning on the architectural (or even any other) level.
This contrasts with our approach, where we enrich the ensemble-based systems with machine-learning techniques.

\section{Conclusion}
\label{sec:conclusion}
In this paper, we have presented ML-DEECo, a novel ensemble-based component model that features abstractions for machine learning (namely the supervised learning) and heuristics related to the assignment of tasks to components (realized by component membership in an ensemble). We illustrated the key abstractions on an example from our past project in Industry 4.0 domain. 

In the future, we aim at featuring abstractions related to unsupervised and reinforcement learning, as they would cover situations when the ground trues cannot be observed at all (note that in this paper we assume that the ground trues cannot be observed at the time the prediction is performed but will be observable at a later point of time).
Another direction for future work is to design a set of common heuristics for ML-DEECo.

\section*{Acknowledgment}

This work has been partially supported by Charles University institutional funding SVV 260588, partially supported by the Czech Science Foundation project 20-24814J, and partially supported by the European Research Council (ERC) under the European Union’s Horizon 2020 research and innovation programme (grant agreement No 810115).

\bibliographystyle{splncs04}
\bibliography{paper}

\end{document}